%% file: neurips_2025.tex
\documentclass{article}

\PassOptionsToPackage{numbers, compress}{natbib}

    \usepackage[preprint]{neurips_2025}

\pdfoutput=1
\usepackage[utf8]{inputenc} 
\usepackage[T1]{fontenc}    
\usepackage{hyperref}       
\usepackage{url}            
\usepackage{booktabs}       
\usepackage{amsfonts}       
\usepackage{nicefrac}       
\usepackage{microtype}      
\usepackage{multirow}
\usepackage{booktabs}
\usepackage[table]{xcolor} 
\usepackage{array}
\usepackage{amsmath} 
\usepackage{graphicx}

\usepackage{listings}    
\usepackage[most]{tcolorbox}
\tcbuselibrary{breakable,skins}
\usepackage{siunitx}
\usepackage{wrapfig}
\usepackage{caption}
\usepackage{adjustbox}

\usepackage[normalem]{ulem}

\definecolor{darkgreen}{RGB}{0, 150, 0} 
\definecolor{darkred}{RGB}{128, 0, 0} 

\title{VCR-Bench: A Comprehensive Evaluation Framework for Video Chain-of-Thought Reasoning}

\author{%
  Yukun Qi$^{1,2}$, Yiming Zhao$^{1,2}$, Yu Zeng$^{1,2}$, Xikun Bao$^{1,2}$,  Wenxuan Huang$^{3}$, Lin Chen{$^{1}$}\thanks{Project leader.},\\
  \textbf{Zehui Chen$^{1}$, Jie Zhao$^{2}$, Zhongang Qi$^{2}$, Feng Zhao{$^{1}$}}\thanks{Corresponding author.} \\
  $^{1}$University of Science and Technology of China \\
  $^{2}$Huawei Noah's Ark Lab \quad $^{3}$East China Normal University\\ \\
  Project Page: \url{https://vlm-reasoning.github.io/VCR-Bench/}
}

\begin{document}

\maketitle

\input{section/0_abstract}
\input{section/1_introduction}
\input{section/2_related_works}
\input{section/3_method}
\input{section/4_Experiment}
\input{section/5_conclusion}

{\small
    \bibliographystyle{abbrv}
    \bibliography{main}
}

\input{appendix/appendix}

\end{document}

%% file: section/0_abstract.tex
\begin{abstract}

The advancement of Chain-of-Thought (CoT) reasoning has significantly enhanced the capabilities of large language models (LLMs) and large vision-language models (LVLMs). However, a rigorous evaluation framework for video CoT reasoning remains absent. Current video benchmarks fail to adequately assess the reasoning process and expose whether failures stem from deficiencies in perception or reasoning capabilities. Therefore, we introduce \textbf{VCR-Bench}, a novel benchmark designed to comprehensively evaluate LVLMs' \textbf{V}ideo \textbf{C}hain-of-Thought \textbf{R}easoning capabilities. VCR-Bench comprises 859 videos spanning a variety of video content and durations, along with 1,034 high-quality question-answer pairs. Each pair is manually annotated with a stepwise CoT rationale, where every step is tagged to indicate its association with the perception or reasoning capabilities. Furthermore, we design seven distinct task dimensions and propose the CoT score to assess the entire CoT process based on the stepwise tagged CoT rationals. Extensive experiments on VCR-Bench highlight substantial limitations in current LVLMs. Even the top-performing model, o1, only achieves a 62.8\% CoT score and an 56.7\% accuracy, while most models score below 40\%. Experiments show most models score lower on perception than reasoning steps, revealing LVLMs' key bottleneck in temporal-spatial information processing for complex video reasoning. A robust positive correlation between the CoT score and accuracy confirms the validity of our evaluation framework and underscores the critical role of CoT reasoning in solving complex video reasoning tasks. We hope VCR-Bench to serve as a standardized evaluation framework and expose the actual drawbacks in complex video reasoning task.

\end{abstract}

%% file: section/1_introduction.tex
\section{Introduction}

The emergence of Chain-of-Thought (CoT) reasoning \cite{wei2022chain} has significantly enhanced the reasoning capability of large language models (LLMs), as evidenced by the recent breakthroughs of DeepSeek-R1 \cite{guo2025deepseek} and OpenAI o1 \cite{o1}. By generating human-like, interpretable reasoning steps, these reasoning models have demonstrated remarkable advantages in solving complex visual tasks. Recently, large vision-language models (LVLMs) \cite{openai2024gpt4o, chen2024sharegpt4v, chen2024we, chen2024sharegpt4video} have achieved groundbreaking progress in multiple visual fields, especially in research on CoT reasoning for video data.

\begin{figure}[t]
  \centering
  \includegraphics[width=0.99\textwidth,keepaspectratio]{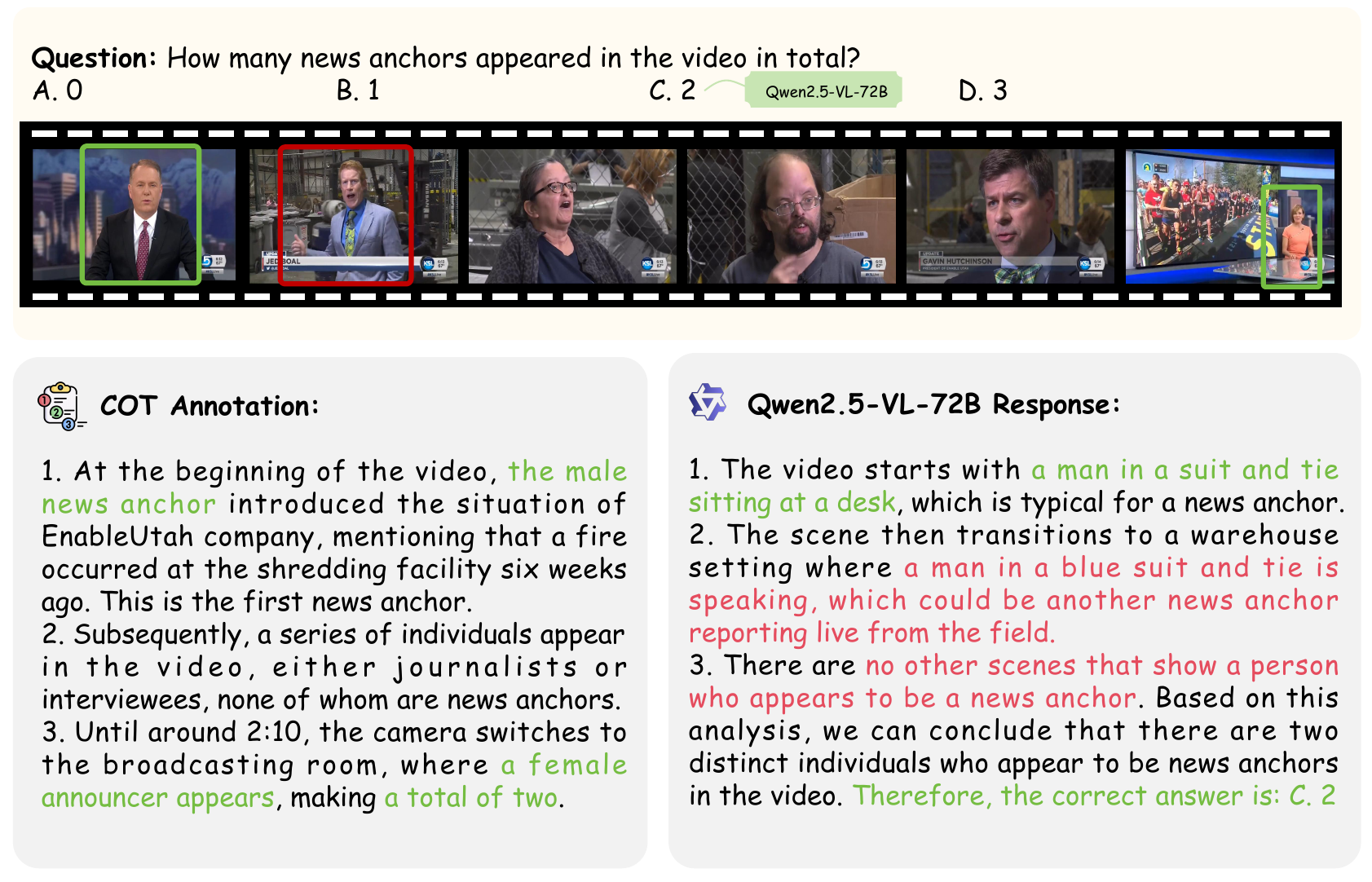}
  \caption{\textbf{Failure case of accuracy-based evaluation.} The video contains  \textcolor{darkgreen}{two news anchors}, but the model missed one while misclassify a \textcolor{darkred}{non-anchor} as an anchor, yet reached the correct answer. This suggests that relying solely on accuracy is insufficient for appropriately evaluating a model's performance under video CoT reasoning.}
  \label{fig:step_wrong}
\end{figure}

However, video understanding field still lacks a scientifically effective evaluation suit for CoT reasoning, with existing benchmarks primarily suffering from the following two shortcomings: First, current video benchmarks \cite{xu2017video, liu2024mmbench, zhou2024mlvu, zhao2025v2p} often lack comprehensive annotations of CoT steps, focusing only on the accuracy of final answers during model evaluation while neglecting the quality of the reasoning process. This evaluation approach makes it difficult to comprehensively evaluate model's actual drawbacks during the CoT reasoning process. As shown in Figure \ref{fig:step_wrong}, the model captures one piece of erroneous information while missing one correct piece during its reasoning process, yet ultimately arrives at the correct final answer. Second, existing video understanding benchmarks \cite{li2023seed, fu2024video} fail to effectively distinguish performance differences in perception and reasoning capabilities. The absence of an effective evaluation suit has become a significant bottleneck that hinders the in-depth development of complex reasoning research in the field of video understanding.

To fill this gap, we propose \textbf{VCR-Bench}, a benchmark specifically designed to evaluate the \textbf{V}ideo \textbf{C}hain-of-Thought \textbf{R}easoning capabilities of LVLMs. We have constructed a multi-dimensional evaluation framework, defining seven distinct task dimensions that comprehensively cover a diverse range of video types and durations. For each data sample, in addition to providing a standard answer, we have meticulously curated detailed and accurate reference stepwise rationals as CoT annotation. All samples underwent rigorous manual annotation and quality control, ultimately resulting in the creation of VCR-Bench, which includes 859 videos and 1,034 high-quality question-answer pairs. We draw on existing work in the field of image understanding \cite{jiang2025mme, chen2024m, thawakar2025llamav} to innovatively design an evaluation framework specifically for assessing generated CoT reasoning steps. This framework first categorizes the CoT steps into visual perception steps and logical reasoning steps, then systematically evaluates the CoT steps across multiple dimensions including recall rate and precision rate to derive the CoT score, thereby providing a basis for comprehensively measuring models' reasoning capabilities.

We conducted a through evaluation of multiple models on our VCR-Bench. The experimental results reveal significant limitations in current models: even the top-performing model, o1 \cite{o1}, achieves only 62.8\% CoT score and 56.7\% accuracy, while most models score below 40\%. This performance gap highlights the notable shortcomings of existing LVLMs in video reasoning tasks and underscores substantial room for improvement. The consistently lower average perception scores compared to reasoning scores indicate that the primary performance bottleneck in current LVLMs for complex video reasoning tasks remains the extraction and comprehension of temporal-spatial information. Further analysis revealed a strong positive correlation between the models' CoT scores and the accuracy. This effectively validates the effectiveness and reliability of our evaluation framework.

In a nutshell, our core contributions are as follows:
\begin{itemize}
    \item To our knowledge, VCR-Bench is the first benchmark specifically designed for video CoT reasoning. Through rigorous manual annotation, we provide detailed reasoning steps for each sample, ensuring data accuracy and reliability while offering the research community a high-quality video reasoning evaluation benchmark.
    \item We have successfully introduced the CoT evaluation framework into the field of video reasoning, assessing the entire reasoning process based on step-by-step annotated CoT rationales, thereby providing an effective approach to measure the video reasoning performance of LVLMs.
    \item Through extensive evaluation experiments, we have validated the effectiveness of our assessment methods and data, while also demonstrating that current LVLMs still exhibit significant limitations in video reasoning, especially in the extraction of temporal-spatial information. Furthermore, our experiments demonstrate a strong correlation between CoT step quality and final answer accuracy.
\end{itemize}

%% file: section/2_related_works.tex
\section{Related Work}
\subsection{LVLMs for Video Understanding}
The rapid advancement of image-based LVLMs \cite{chen2024sharegpt4video, lin2023video, ye2024mplug, maaz2023video} has significantly boosted video understanding and question answering capabilities, revitalizing AI research. Early attempts like VideoChat and Video-ChatGPT \cite{maaz2023video} paved the way for recent advancements such as CogVLM2-Video \cite{hong2024cogvlm2}, InternVL2 \cite{chen2024internvl, chen2024far}, and LLaVA-Video \cite{zhang2024video}, which process videos as image sequences by leveraging powerful image comprehension. To address the computational challenges of high frame rates and long videos, techniques like QFormer-based feature extraction in InternVideo2 \cite{wang2024internvideo2} and Video-LLaMA \cite{zhang2023video}, and adaptive pooling in PLLaVA \cite{xu2024pllava} have been developed. With the enhancement of model capabilities and the increasing complexity of tasks, the strong reasoning and thinking abilities of LVLMs in the field of video understanding are receiving growing attention.

\subsection{Video Understanding Benchmarks}
Traditional video understanding benchmarks focus on evaluating specific model capabilities in particular scenarios. For example, MSRVTT-QA \cite{xu2017video}, ActivityNet-QA \cite{yu2019activitynet}, and NExT-QA \cite{xiao2021next} test basic action recognition and video question answering, while MMBench \cite{xu2023mmbench}, SEED-Bench \cite{li2023seed}, and MVBench \cite{li2024mvbench} assess short video clips. Benchmarks like LongVideoBench \cite{wu2024longvideobench}, Video-MME \cite{fu2024video}, and LVBench \cite{wang2024lvbench} provide longer videos and more diverse tasks. Latest work, such as V2P-Bench \cite{zhao2025v2p}, has constructed a set of data based on visual prompts by simulating human-computer interactions. However, these tasks are generally simple and do not require complex reasoning from models. Recently, there has been growing interest in video CoT reasoning tasks. VideoEspresso \cite{han2024videoespresso} uses keyframe captions for complex scene reasoning, MMVU \cite{zhao2025mmvu} introduces annotated educational video reasoning questions, and VideoMMMU \cite{hu2025video} focuses on knowledge reasoning from subject explanation videos. While these efforts aim to measure video CoT reasoning, their scenarios are limited, and they primarily evaluate final results rather than the reasoning process itself.

\subsection{Reasoning Evaluation}
In the multimodal domain, research on evaluating reasoning processes remains relatively scarce and is primarily focused on the image domain. Early efforts to assess reasoning capabilities were mainly concentrated in scientific fields, such as MathVista \cite{lu2023mathvista}, MathVerse \cite{zhang2024mathverse}, and OlympiadBench \cite{he2024olympiadbench}, which are limited to overly specific scenarios. Recent works have extended the evaluation of reasoning processes to the general image domain. For instance, M³CoT \cite{chen2024m} and SciVerse \cite{sciverse} incorporate commonsense tasks, scientific reasoning, and knowledge-based assessment into multimodal benchmarks. However, these works still lack comprehensive evaluation of the reasoning process. LlamaV-o1 \cite{thawakar2025llamav} constructs a multi-dimensional evaluation framework to meticulously assess image reasoning processes, while MME-CoT \cite{jiang2025mme} achieves promising results in process evaluation within the image domain by matching output steps with annotated steps and establishing an $F_1$ score calculation criterion. These methodologies can be adapted and applied to the field of video reasoning.

%% file: section/3_method.tex
\section{VCR-Bench}

\subsection{Dataset Curation}
\begin{figure*}[!t]
\centering
\begin{minipage}[c]{0.49\textwidth}
\includegraphics[width=\columnwidth]{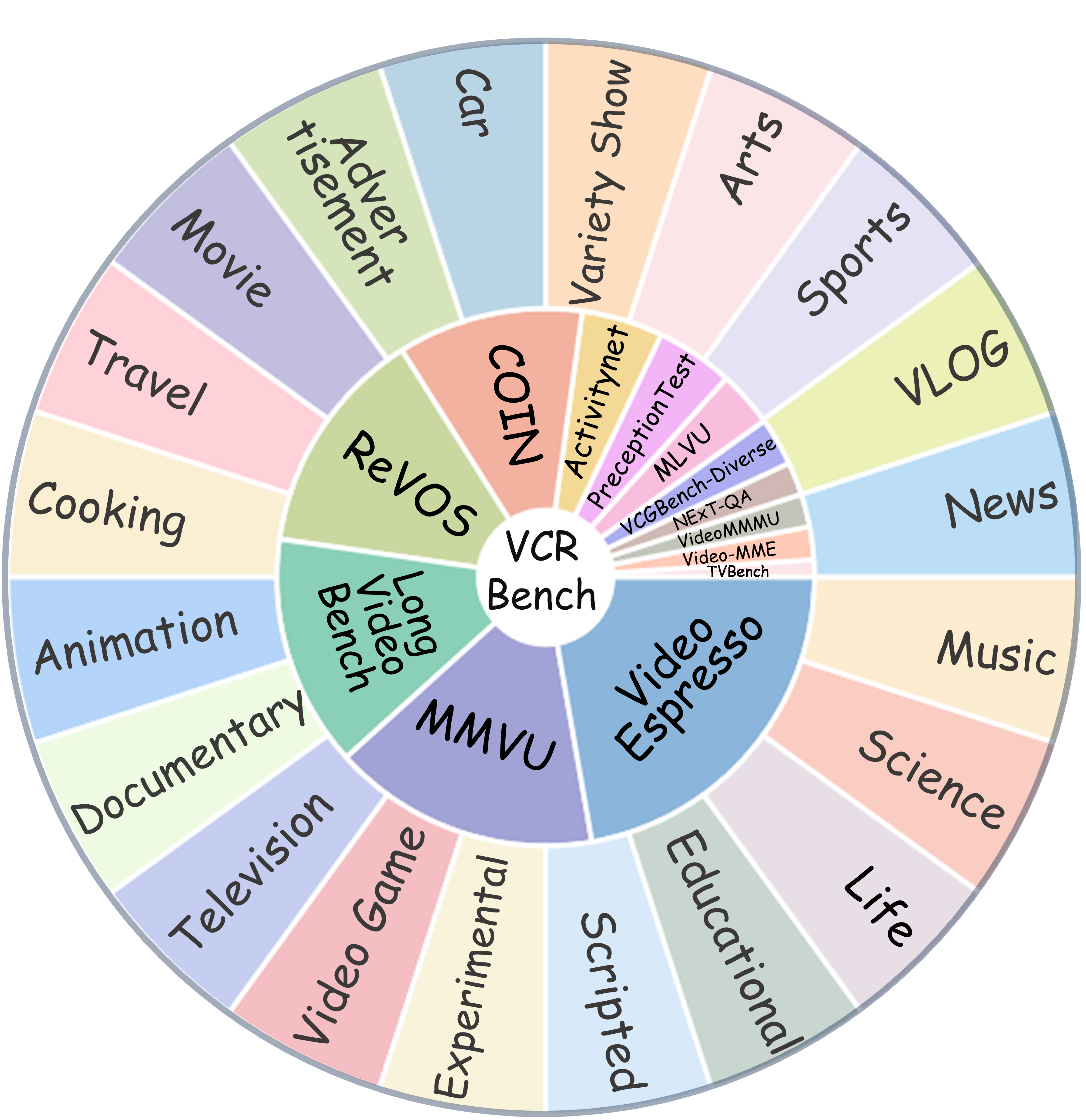}
\caption{\textbf{Video source and categories.}} 
\label{pie}
\end{minipage} \hspace{3pt}   
\begin{minipage}[c]{0.49\textwidth}
\captionof{table}{\textbf{Key Statistics of VCR-Bench.}}
\small
\centering
\begin{adjustbox}{width=0.83\textwidth}
\begin{tabular}{lc}
\toprule
\textbf{Statistic} & \textbf{Number} \\
\midrule
Total Videos & 859\\
\quad -  Short Videos ($\leq1$ min) &418 (48.7\%)  \\
\quad -  Medium Videos ($1\sim5$ min) &293 (34.1\%)  \\
\quad -  Long Videos ($>5$ min) &148 (17.2\%)   \\
\midrule
Total Questions & 1034  \\
\quad - Dimensions \\
\quad \quad  Fundamental Temporal Reasoning &159 (15.4\%)  \\
\quad \quad  Video Temporal Counting &161 (15.6\%)  \\
\quad \quad  Video Temporal Grounding &143 (13.8\%)  \\
\quad \quad  Video Knowledge Reasoning &153 (14.8\%)  \\
\quad \quad  Temporal Spatial Reasoning &135 (13.1\%)  \\
\quad \quad  Video Plot Analysis &139 (13.4\%)  \\
\quad \quad  Temporal Spatial Grounding &144 (13.9\%)  \\
\quad - Types \\
\quad \quad Multiple-choice &510 (49.3\%) \\
\quad \quad Open-ended &524 (50.7\%) \\
\midrule
Total Reference Reasoning Steps &4078  \\
\quad - Visual Perception Steps &2789 (68.4\%)  \\
\quad - Logical Reasoning Steps &1289 (31.6\%)  \\
Reasoning Steps per Sample (avg/max) &3.9/12  \\
Reasoning Step Word Count (avg/max) &27.0/129  \\
\midrule
Question Word Count  (avg/max) &22.1/161  \\
Answer Word Count (avg/max) &3.5/49  \\
\bottomrule
\end{tabular}
\end{adjustbox}
\label{table:statistics}
\end{minipage}
\vspace{-0.2cm}
\end{figure*}

As shown in Figure \ref{pie}, to ensure the diversity of video data and the richness of sample information, we curated the VCR-Bench by selecting and integrating data from multiple existing video benchmarks. These include datasets focused on video perception and comprehension, such as Perception Test \cite{patraucean2024PerceptionTest}, NExTVideo \cite{xiao2021next}, TVbench \cite{cores2024tvbench}, MLVU \cite{zhou2024mlvu}, VCGBench-Diverse \cite{Maaz2024VideoGPT+} and COIN \cite{tang2019coin}; datasets targeting subject knowledge understanding and reasoning, such as videoMMMU \cite{hu2025video} and MMVU \cite{zhao2025mmvu}; datasets emphasizing long-form video understanding, including Video-MME \cite{fu2024video} and LongVideoBench \cite{wu2024longvideobench}; datasets specialized in video temporal localization and analysis, such as ActivityNet Captions \cite{krishna2017dense} and ReVOS Videos \cite{yan2024visa}; as well as datasets dedicated to video scene reasoning, exemplified by VideoEspresso \cite{han2024videoespresso}, among others.

\subsubsection{Task Definition}
\label{sec:Task Definition}
\begin{figure}[t]
  \centering
  \includegraphics[width=0.99\textwidth, keepaspectratio]{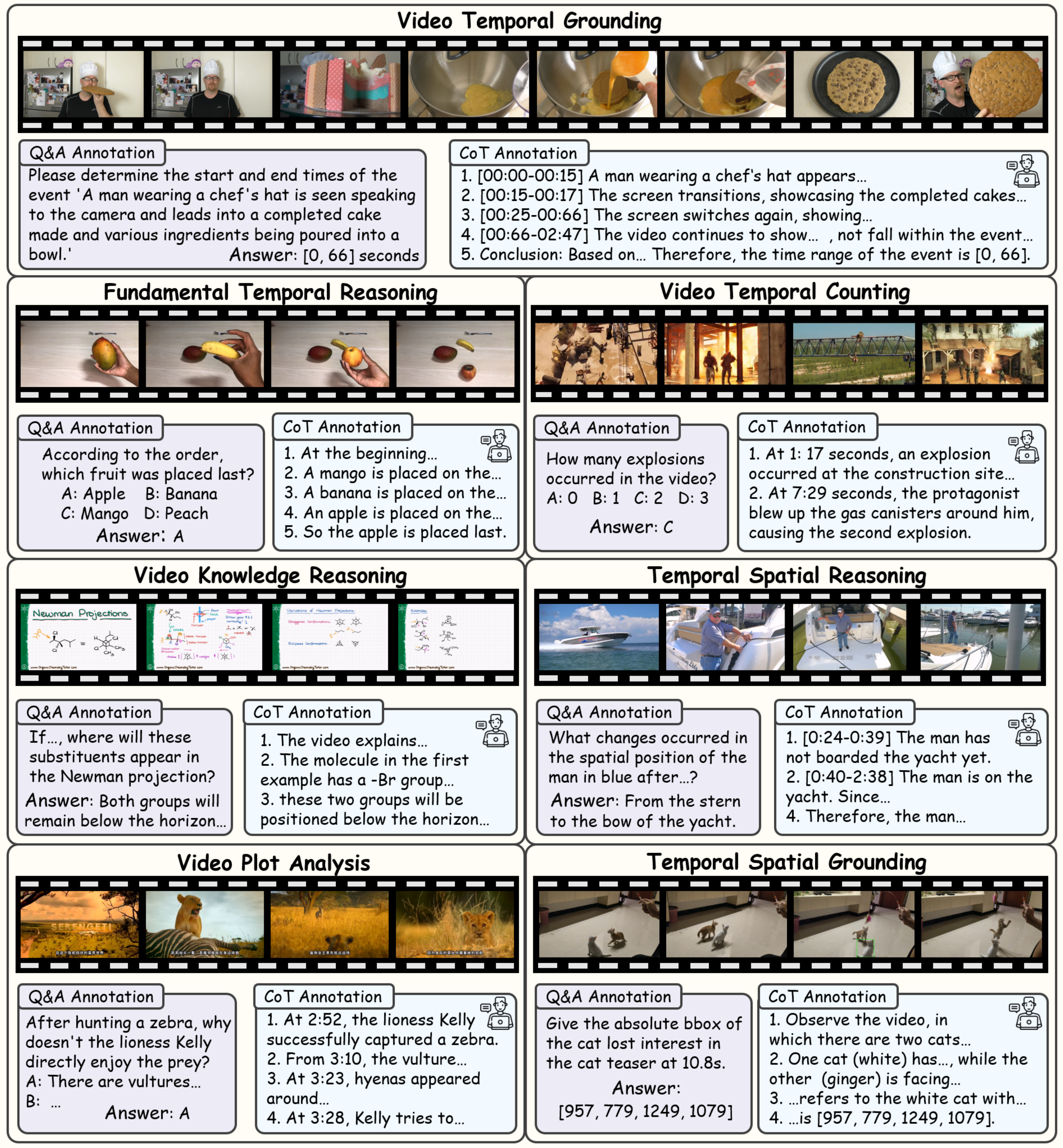}
  \caption{\textbf{Cases across dimensions.} VCR-Bench encompasses seven distinct task dimensions spanning multiple competency levels, including spatiotemporal perception, logical reasoning, and knowledge-based analysis.}
  \label{fig:cases_dim}
\end{figure}

To comprehensively evaluate the differences in LVLMs' capabilities for video Chain-of-Thought (CoT) reasoning from multiple perspectives, we define seven distinct dimensions of task categories, as illustrated in Figure \ref{fig:cases_dim}. These dimensions encompass various aspects such as spatiotemporal perception, logical reasoning, and knowledge-based analysis. The specific task types are as follows:

~~$\bullet$ \textbf{Fundamental Temporal Reasoning (FTR):} FTR task represents a basic temporal reasoning problem, requiring the model to develop a deep understanding of the temporal order and to analyze and compare the sequence in which events or actions occur.

~~$\bullet$ \textbf{Video Temporal Counting (VTC):}  VTC task requires the model to calculate the frequency of events or actions and to perceive the number of occurrences of specific objects.

~~$\bullet$ \textbf{Video Temporal Grounding (VTG):} VTG task requires the model to locate the specific moment or time interval corresponding to a given action or event.

~~$\bullet$ \textbf{Video Knowledge Reasoning (VKR):} VKR task requires the model to extract specific knowledge-related information from the video and apply domain-specific logical reasoning to solve targeted problems.

~~$\bullet$ \textbf{Temporal Spatial Reasoning (TSR):} TSR task focuses on the spatial position changes of characters within the video, including their movement trajectories and specific locations.

~~$\bullet$ \textbf{Video Plot Analysis (VPA):} VPA task requires the model to understand the narrative logic of the video and provide explanations for specific events that occur within the plot.

~~$\bullet$ \textbf{Temporal Spatial Grounding (TSG):} TSG task requires the model to locate the spatial position of a corresponding object within a specified temporal sequence.

\subsubsection{Data Annotation and Review}
\label{sec:Data Annotation and Review}
To enable CoT evaluation, we provide questions, answers, and CoT annotations (reference reasoning steps) for all data. These reference steps represent the essential reasoning path to derive correct answers. Our annotation pipeline combines automated generation (using Gemini 2.0 \cite{pichai2024introducing}) followed by human verification. This ensures both diversity and accuracy. Each sample's reasoning steps form an ordered set $\mathcal{R} = \{ r_{1}, r_{2}, ..., r_{N} \}$ of $N$ atomic sub-steps, designed to facilitate granular evaluation.

\subsubsection{Data Analysis}
After data annotation and verification, we have ultimately constructed a dataset comprising 859 videos and 1034 question-answer pairs. As shown in Table \ref{table:statistics}, our video dataset encompasses a wide range of different scenarios, including indoor daily life, sports competitions, outdoor nature, and urban architecture. It covers multiple categories such as personal photography, documentaries, films and television, educational videos, and news reports. The duration of the videos ranges from less than one minute to over 30 minutes, ensuring rich diversity in content and high density of informational cues. Meanwhile, our question-answer pair data achieves a rough balance across seven different dimensions, ensuring the richness and balance of the benchmark tasks.

\begin{figure}[t]
  \centering
  \includegraphics[width=0.99\textwidth, keepaspectratio]{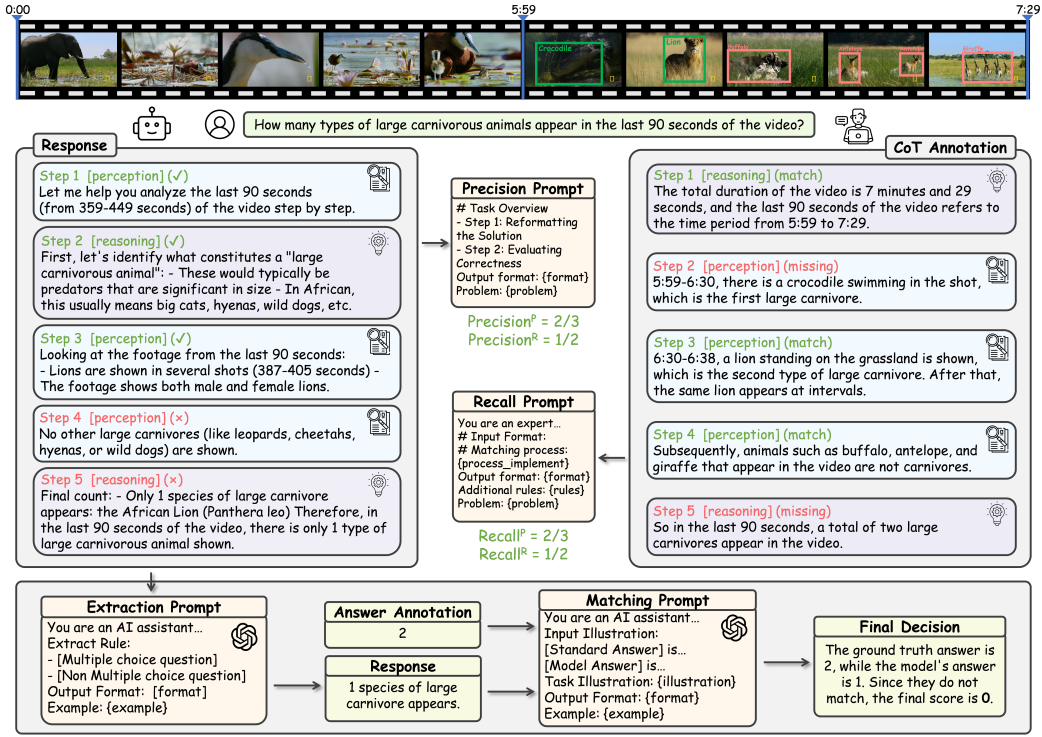}
  \caption{\textbf{Overview of VCR-Bench.} For each sample, we provide detailed CoT annotations. During evaluation, we decompose model responses into reasoning steps and match them with reference CoT to compute recall/precision. Final answers are extracted and compared against ground-truth.}
  \label{fig:main}
\end{figure}

\subsection{CoT Evaluation Strategy}
Current video understanding benchmarks primarily evaluate the correctness of models' final answers while neglecting intermediate CoT reasoning steps. This evaluation approach fails to provide a comprehensive assessment of models' reasoning capabilities. When addressing complex problems, models must perform multiple cognitive operations including perception and reasoning - evaluating only the final answers cannot reveal their actual shortcomings. As shown in Figure \ref{fig:main}, to address this limitation, our proposed VCR-Bench incorporates two additional evaluation components alongside conventional final-answer assessment: CoT Reasoning Deconstruction and CoT Quality Evaluation.

\subsubsection{CoT Reasoning Deconstruction}
The reasoning process of LVLMs involves multiple distinct operations, reflecting diverse capabilities. To systematically evaluate model performance across these competencies, we propose CoT Reasoning Deconstruction, which breaks down the process into two core dimensions:

\textbf{Visual Perception} assesses the model’s ability to extract spatiotemporal information (\textit{e.g.}, actions, object locations) from videos—the foundational skill for vision tasks.

\textbf{Logical Reasoning} evaluates the model’s capacity to derive conclusions from perceived information, critical for complex problem-solving.

Formally, we represent reference reasoning steps as: $\mathcal{R} = \mathcal{R}_{p} \cup \mathcal{R}_{r}$, where the $\mathcal{R}_{p}$ and $\mathcal{R}_{r}$ denote \textbf{perception} and \textbf{reasoning} subprocesses, respectively.

\subsubsection{CoT Quality Evaluation}
As described in Section \ref{sec:Data Annotation and Review}, the question-answer pairs in the VCR-Bench provide accurate and concise reference reasoning steps $\mathcal{R}$. The core of evaluating the model's reasoning content is to establish a matching relationship between the model's reasoning steps $\mathcal{S}$ and the reference reasoning steps $\mathcal{R}$, to determine the correctness of the model's reasoning. To this end, we use GPT4o \cite{openai2024gpt4o} to decompose the model's reasoning content into $K$ independent and structurally similar sub-steps, and categorize them into two sub-processes, as shown in Eq. \ref{eq:2}.

\begin{equation}
\label{eq:2}
\mathcal{S}  =  \mathcal{S}_{p} \cup \mathcal{S}_{r}  = \left\{ s_{1}, s_{2}, s_{3}, \dots, s_{K} \right\}
\end{equation}

Then, we evaluate the reasoning process of the model under test based on the following metrics:

\textbf{Recall.} For each sub-step $r_{i}$ in $\mathcal{R}$, we prompt GPT4o to evaluate whether the corresponding content of $r_{i}$ also appears in $\mathcal{S}$. If the same content appears in $\mathcal{S}$ and is entirely correct — including accurate temporal localization, correct entity recognition, and consistent logical reasoning — then $r_{i}$ is considered matched and denoted as $r_{i}^{\text{match}}$. The set of all matched sub-steps is denoted as $\mathcal{R}^{\text{match}}$, and $\mathcal{R}^{\text{match}} = \mathcal{R}_{p}^{\text{match}} \cup \mathcal{R}_{r}^{\text{match}}$. The $Recall$ can be calculated as shown in the following Eq. \ref{eq:recall}.

\begin{equation}
\label{eq:recall}
Recall_{p} = \frac{\left| \mathcal{R}^{\text{match}}_{p} \right|}{\left| \mathcal{R}_{p} \right|}, Recall_{r} = \frac{\left| \mathcal{R}^{\text{match}}_{r} \right|}{\left| \mathcal{R}_{r} \right|},
Recall = \frac{\left| \mathcal{R}^{\text{match}} \right|}{\left| \mathcal{R} \right|}
\end{equation}

The $Recall$ metric comprehensively evaluates the reasoning process by comparing the model's output with the reference solution's key reasoning steps. This metric not only verifies answer correctness but also rigorously examines the logical robustness of the reasoning, effectively eliminating random guessing scenarios, thereby enabling in-depth assessment of the model's reasoning capabilities.

\textbf{Precision.} For each sub-step $s_{j}$ in $\mathcal{S}$, we prompt GPT4o to evaluate based on the content of $\mathcal{R}$ whether $s_{j}$ is accurate. If $s_{j}$ matches and is correct according to the content in $\mathcal{R}$, it is considered a correct step, denoted as $s_{j}^{\text{correct}}$. If $s_{j}$ does not match or contradicts the content in $\mathcal{R}$, such as errors in the temporal localization of key events, or mistakes in causal reasoning, it is considered an incorrect step, denoted as $s_{j}^{\text{incorrect}}$. If $s_{j}$ does not appear in $\mathcal{R}$, or it is impossible to determine whether $s_{j}$ is correct based on the content in $\mathcal{R}$, it is considered an irrelevant reasoning step in solving the problem, denoted as $s_{j}^{\text{irrelevant}}$. The set of correct steps and incorrect steps are denoted as $\mathcal{S}^{\text{correct}}$ and $\mathcal{S}^{\text{incorrect}}$. Similarly, both $\mathcal{S}^{\text{correct}}$ and $\mathcal{S}^{\text{incorrect}}$ can be further decomposed into the form as shown in \ref{eq:P1}.

\begin{equation}
\label{eq:P1}
\mathcal{S}^{\text{correct}} = \mathcal{S}_{p}^{\text{correct}} \cup \mathcal{S}_{r}^{\text{correct}},
\mathcal{S}^{\text{incorrect}} = \mathcal{S}_{p}^{\text{incorrect}} \cup \mathcal{S}_{r}^{\text{incorrect}}
\end{equation}

Accordingly, the $Precision$ can be calculated as shown in the following Eq. \ref{eq:Precision1} and Eq. \ref{eq:Precision2}.

\begin{align}
\label{eq:Precision1}
Precision_p &= \frac{\left| \mathcal{S}^{\text{correct}}_{p}\right|}{\left| \mathcal{S}^{\text{correct}}_{p} \cup \mathcal{S}^{\text{incorrect}}_{p} \right|}, Precision_r = \frac{\left| \mathcal{S}^{\text{correct}}_{r}\right|}{\left| \mathcal{S}^{\text{correct}}_{r} \cup \mathcal{S}^{\text{incorrect}}_{r} \right|}
\end{align}

\begin{equation}
\label{eq:Precision2}
    Precision  = \frac{\left| \mathcal{S}^{\text{correct}}\right|}{\left| \mathcal{S}^{\text{correct}} \cup \mathcal{S}^{\text{incorrect}} \right|}
\end{equation}

The $Precision$ metrics evaluate the model's output reasoning steps, assessing whether each step is truly reliable and closely related to the answer. By combining $Precision$ and $Recall$ metrics, we can calculate the model's output $F_1$ score as shown in Equation \ref{eq:6} to serve as the final CoT score, thereby enabling more reliable and comprehensive evaluation of the model's CoT response quality.

\begin{equation}
\label{eq:6}
    F_1 = 2 \cdot \frac{Precision \cdot Recall}{Precision+Recall}
\end{equation}

\subsection{Accuracy Evaluation Strategy}
For the accuracy evaluation of the model's final results, we adopted the following approach: First, we used the GPT4o \cite{openai2024gpt4o} model to extract the final answer from the model's output CoT steps. For general question-answering tasks, GPT4o \cite{openai2024gpt4o} was employed to evaluate whether the extracted final answer was correct based on human-annotated reference answers. For more specialized tasks such as VTG and TSG, we calculated the Intersection over Union (IoU) between the extracted final answer and the reference answer. Samples with an IoU greater than a specified threshold were judged as correct. The IoU threshold was set to 0.7 for VTG tasks and 0.5 for TSG tasks.

\input{table/dim}

%% file: table/dim.tex
\begin{table}[t]
\centering
\caption{\textbf{CoT Evaluation Results for Different Models in VCR-Bench.} The best results are \textbf{bold} and the second-best are \underline{underlined}. The $F_1$ represents the final CoT score.}
\begin{tabular}{>{\raggedright\arraybackslash}p{3.4cm}|p{0.65cm}p{0.65cm}p{0.65cm}|p{0.65cm}p{0.65cm}p{0.65cm}|p{0.65cm}p{0.65cm}p{0.65cm}}
\toprule
\multirow{2}{*}{Model} & \multicolumn{3}{c|}{Perception} & \multicolumn{3}{c|}{Reasoning} & \multicolumn{3}{c}{Avg} \\
\cmidrule(lr){2-4} \cmidrule(lr){5-7} \cmidrule(lr){8-10}
& $Rec$ & $Pre$ & $F_1$ & $Rec$ & $Pre$ & $F_1$ & $Rec$ & $Pre$ & $F_1$ \\
\midrule
\rowcolor{gray!20} \multicolumn{10}{l}{\textit{Closed-Source Models}} \\ 
Gemini-2.0-Flash &\underline{52.1} &66.6 &\underline{58.5} &\underline{57.4} &\underline{64.6} &\underline{60.8} &\underline{54.0} &62.1 &\underline{57.7} \\
Gemini-1.5-Pro &47.1 &57.8 &51.9 &54.8 &54.3 &54.5 &49.4 &54.3 &51.7 \\
o1 &\textbf{52.4} &70.0 &\textbf{59.9} &\textbf{66.6} &\textbf{71.4} &\textbf{68.9} &\textbf{56.9} &\textbf{70.1} &\textbf{62.8}\\
GPT-4o &51.4 &61.0 &55.8 &55.3 &52.4 &53.8 &52.7 &56.9 &54.7 \\
Claude 3.5 Sonnet &47.7 &58.1 &52.4 &49.1 &47.5 &48.3 &47.6 &53.6 &50.4 \\
\rowcolor{gray!20} \multicolumn{10}{l}{\textit{Open-Source Models}} \\ 
InternVL2.5-8B &16.1 &52.6 &24.6 &33.0 &36.9 &34.8 &22.1 &38.2 &28.0 \\
InternVL2.5-78B &18.7 &74.1 &29.9 &35.2 &53.9 &42.6 &23.9 &56.8 &33.7 \\
VideoLLaMA3-7B &20.2 &52.2 &29.1 &39.1 &39.9 &39.5 &26.6 &40.1 &32.0 \\
LLaVA-OneVision-7B &10.1 &92.3 &18.3 &28.7 &51.2 &36.8 &16.7 &55.1 &25.6 \\
LLaVA-OneVision-72B &14.1 &94.7 &24.5 &35.5 &58.3 &44.1 &20.8 &61.5 &31.1 \\
mPLUG-Owl3-7B &6.0 &86.5 &11.1 &20.7 &43.7 &28.1 &10.4 &45.4 &17.0 \\
MiniCPM-o2.6-8B &27.5 &49.4 &35.3 &34.6 &35.0 &34.8 &29.9 &38.7 &33.8 \\
Llama-3.2-11B-Vision &2.1 &86.4 &4.2 &6.8 &52.5 &12.0 &3.6 &52.5 &6.8 \\
Qwen2.5-VL-7B &31.7 &53.4 &39.8 &34.7 &37.4 &36.0 &33.4 &44.6 &38.2 \\
Qwen2.5-VL-72B &46.2 &60.2 &52.3 &47.4 &46.1 &46.7 &47.5 &53.8 &50.5 \\
LLaVA-Video-7B &11.1&\underline{95.7} &19.9 &33.1 &52.0 &40.4 &18.1 &56.4 &27.3 \\
LLaVA-Video-72B &15.6 &95.3 &26.9 &39.8 &57.1 &46.9 &23.2 &60.6 &33.6 \\
Aria-25B &18.5 &68.6 &29.1 &36.2 &52.3 &42.8 &23.9 &56.0 &33.5 \\
InternVideo2.5-8B &6.9 &\textbf{98.4} &12.9 &26.1 &61.3 &36.6 &12.6 &\underline{66.0} &21.2 \\
\bottomrule
\end{tabular}
\label{tab:metrics}
\end{table}

%% file: section/4_Experiment.tex
\section{Experiments}

\subsection{Experiment Setup}
\textbf{Evaluation Models.}
To thoroughly evaluate the effectiveness of VCR-Bench, we conducted assessments on multiple models. These include mainstream and powerful closed-source models such as Gemini (1.5 Pro, 2.0 Flash) \cite{team2024gemini1.5, pichai2024introducing}, GPT4o \cite{openai2024gpt4o}, o1 \cite{o1}, and Claude 3.5 \cite{anthropic2024claude3}, as well as commonly used open-source models like InternVL2.5 (8B, 78B) \cite{chen2024internvl, chen2024far, chen2024expanding}, VideoLLaMA3 (7B) \cite{zhang2025videollama}, LLaVA-OneVision (7B, 72B) \cite{li2024llava}, mPLUG-Owl3 (7B) \cite{ye2024mplug}, MiniCPM-o2.6 (7B) \cite{yao2024minicpm}, Llama-3.2-Vision (11B) \cite{llama3modelcard}, Qwen2.5-VL (7B, 72B) \cite{Qwen2.5-VL}, LLaVA-Video (7B, 72B) \cite{zhang2024video}, Aria (25B) \cite{li2024aria}, and InternVideo2.5 (8B) \cite{wang2025internvideo}. This essentially covers all the mainstream LVLMs currently available.

\textbf{Implementation Details.}
For models supporting direct video input, such as Gemini \cite{team2024gemini1.5, pichai2024introducing}, we processed the videos directly. For models currently without native video support (e.g., GPT-4o \cite{openai2024gpt4o}), we extracted 64 frames per video with corresponding timestamp annotations, using multi-image input for evaluation. All other model parameters strictly followed official specifications. During inference, all models were required to answer questions step-by-step using our defined CoT prompt: "\textit{Please provide a step-by-step solution to the given question}." All other prompts used during evaluation are provided in the Appendix \ref{appendix: prompt}.

\subsection{CoT Evaluation Results}
We first evaluated the output CoT steps of each model, and the experimental results are shown in Table \ref{tab:metrics}. From the results, it can be observed that the quality of output CoT varies significantly across different models, and the overall CoT scores are not particularly high. Among them, the o1 \cite{o1} model, which focuses on strong reasoning capabilities, achieved the highest CoT scores in both the Perception and Reasoning dimensions, with a comprehensive CoT score of 62.8, the highest among all models. Further analysis of the results leads us to the following conclusions:

\input{table/acc}
\textbf{Closed-source models and large-scale parameter models possess stronger reasoning capabilities.}
As shown in the results of Table \ref{tab:metrics}, the CoT evaluation CoT scores of common closed-source models are generally higher than those of open-source models. Additionally, for the same open-source model with different parameter sizes, such as Qwen2.5-VL 7B and 72B \cite{Qwen2.5-VL}, the model with larger parameters achieves a higher CoT score. This reflects that video CoT reasoning places high demands on the overall performance of LVLMs, and only models with larger parameters can ensure better step-by-step analysis and reasoning capabilities.

\textbf{A more common issue that models encounter during multi-step reasoning is omission rather than inaccuracy.}
Experimental results demonstrate that most models achieve higher precision scores than recall scores. For some models with weaker CoT reasoning capabilities (\textit{e.g.}, LLaVA-Video-7B \cite{zhang2024video}), their outputs typically contain only one or two reasoning steps, which further widens this performance gap. This indicates that while the majority of the reasoning steps generated by the models are accurate and valid, there still exists significant omission of critical reasoning steps.

\textbf{The logical reasoning performance of the models is generally stronger than their visual perception performance.} 
The models' logical reasoning performance is generally stronger than their visual perception performance. Quantitative analysis of the table results demonstrates that their average reasoning capability (mean CoT score 42.5) surpasses their average perception ability (mean CoT score 33.5), with this performance gap being particularly pronounced among open-source models exhibiting performance deviations. This reveals that the current performance bottleneck of LVLMs in complex video reasoning tasks primarily lies in visual perception information extraction and comprehension.

\subsection{Accuracy Evaluation Results}
As shown in Table \ref{tab:acc}, we evaluated the final answer accuracy of all models across different dimensions. Combined with the results from Table \ref{tab:metrics}, we can draw the following conclusions:

\textbf{The CoT evaluation results are highly positively correlated with the final answer evaluation results.}
As shown in Figure~\ref{fig:correlation}, the experimental results demonstrate a strong positive correlation (r=0.89) between models' CoT reasoning quality and final answer accuracy. This robust relationship confirms that effective CoT reasoning is critical for successful video question answering, with higher-quality CoT steps consistently leading to more accurate final responses. 

\textbf{Models with stronger instruction-following capabilities can achieve relatively higher CoT scores.}
A closer examination of Figure~\ref{fig:correlation} reveals that some models exhibit relatively high accuracy but low CoT scores, such as LLaVA-Video-7B \cite{zhang2024video} and LLaVA-OneVision-7B \cite{li2024llava}. These models generally struggle to properly follow CoT instructions—even when provided with CoT prompts, their outputs remain overly concise, and their reasoning processes are insufficiently detailed, resulting in lower CoT scores. In contrast, models like Qwen2.5-VL \cite{Qwen2.5-VL}, which demonstrate stronger instruction-following capabilities, produce more comprehensive reasoning chains, thus achieving comparatively higher CoT scores.

\textbf{The spatiotemporal grounding capabilities of the models are generally weak.}
The TSG task proves exceptionally challenging, with even the top model (Gemini-2.0-Flash \cite{pichai2024introducing}) achieving merely 4.2\% accuracy, while many models fail completely. This stems from the task's unique demands: (1) combined spatiotemporal reasoning (temporal localization + coordinate output), and (2) current models' fundamental limitations in extracting precise spatial coordinates from video data. For concrete examples, please refer to Figure \ref{fig:error case2} in the Appendix \ref{appendix: error analysis}.

\subsection{More Evaluation Results}

\input{table/merge}

\textbf{Accuracy Evaluation Results for Different Durations.}
We also statistically analyzed the model's performance across videos of different durations, as shown in Table \ref{tab:duration}. The results indicate that the model generally achieves better performance on medium-length videos. In comparison, long videos contain more complex temporal information and richer content, which poses greater challenges for the model's comprehension. As for short videos, since our dataset is primarily based on manual annotations and corrections, human annotators tend to find them easier to understand and are thus able to produce more in-depth and sophisticated annotations. Meanwhile, the model shows significant deficiencies in the TSG dimension, which mainly consists of short videos. This partially contributes to its weaker performance on short-form content.

\textbf{Accuracy Evaluation Results under Different Settings.}
To further validate the rationality of VCR-Bench, we conducted experiments under different settings, including: text-only input without video, text plus a single frame extracted from video, and full text plus video with direct answering (without CoT), compared with our standard setup of full text plus video with CoT answering. As shown in Table \ref{tab:setting}, both the text-only and single-frame input settings lead to significant performance degradation, indicating that our question-answer data highly depend on video content and temporal information. Meanwhile, for stronger closed-source models, using CoT prompting results in higher accuracy than direct answering, whereas the opposite is true for weaker open-source models. This demonstrates that effective CoT reasoning heavily relies on the model's overall capability—only models with sufficiently strong reasoning skills can fully benefit from CoT.

%% file: table/acc.tex
\begin{table}[t]
\centering
\caption{\textbf{Accuracy Evaluation Results for Different Models in VCR-Bench.} The best results are \textbf{bold} and the second-best are \underline{underlined}.} 
\setlength{\tabcolsep}{6pt} 
\begin{tabular}{>{\raggedright\arraybackslash}p{4.5cm}|*{8}{c}} 
\toprule
Model & FTR & VTC & VTG & VKR & TSR & VPA & TSG & Avg \\
\hline
\rowcolor{gray!20} \multicolumn{9}{l}{\textit{Closed-Source Models}} \\ 
Gemini-2.0-Flash &\underline{66.2} &\underline{51.2} &\textbf{62.0} &64.4 &\underline{54.1} &\underline{58.1} &\textbf{4.2} &\underline{51.7} \\
Gemini-1.5-Pro &55.1 &45.3 &52.9 &62.0 &45.0 &45.6 &0.7 &44.0 \\
o1 &\textbf{66.7} &\textbf{52.2} &\underline{56.9} &\textbf{74.3} &\textbf{61.0} &\textbf{60.2} &0.0 &\textbf{56.7} \\
GPT-4o &54.7 &49.1 &44.8 &\underline{68.6} &48.9 &57.6 &\underline{2.8} &46.9 \\
Claude 3.5 Sonnet &45.3 &46.3 &34.3 &64.2 &44.0 &49.3 &0.7 &41.0 \\
\midrule
\rowcolor{gray!20} \multicolumn{9}{l}{\textit{Open-Source Models}} \\ 
InternVL2.5-8B &32.7 &29.8 &11.9 &33.3 &25.9 &30.9 &0.7 &23.9 \\
InternVL2.5-78B &40.9 &39.8 &9.8 &52.9 &29.6 &39.6 &0.0 &30.9 \\
VideoLLaMA3-7B &44.7 &36.6 &24.5 &43.1 &36.3 &39.6 &0.7 &32.5 \\
LLaVA-OneVision-7B &35.8 &34.8 &24.5 &39.9 &37.8 &41.0 &0.0 &30.7 \\
LLaVA-OneVision-72B &47.8 &42.2 &25.9 &52.3 &45.9 &38.1 &0.0 &36.4 \\
mPLUG-Owl3-7B &13.2 &6.2 &2.8 &5.9 &15.6 &7.2 &0.0 &7.3 \\
MiniCPM-o2.6-8B &31.4 &30.4 &12.6 &43.8 &30.4 &38.1 &0.0 &26.9 \\
Llama-3.2-11B-Vision &4.4 &4.3 &7.0 &6.5 &6.7 &5.8 &0.0 &4.9 \\
Qwen2.5-VL-7B &37.1 &26.7 &29.4 &47.1 &34.8 &36.0 &0.7 &30.4 \\
Qwen2.5-VL-72B &45.0 &39.9 &34.1 &56.2 &38.1 &48.9 &2.1 &37.9 \\
LLaVA-Video-7B &47.2 &36.6 &18.9 &41.8 &40.7 &40.3 &0.0 &32.5 \\
LLaVA-Video-72B &49.7 &49.1 &17.5 &49.7 &43.7 &43.2 &0.0 &36.6 \\
Aria-25B &45.3 &45.0 &33.6 &56.2 &43.7 &38.8 &\underline{2.8} &38.2 \\
InternVideo2.5-8B &40.9 &43.5 &14.0 &41.2 &48.1 &41.7 &0.0 &33.0 \\
\bottomrule
\end{tabular}
\label{tab:acc}
\end{table}

%% file: table/merge.tex
\begin{figure}[t]
\centering
\begin{minipage}{0.48\textwidth}
\small
\centering
\centering
\captionof{table}{Accuracy Evaluation Results for Different Durations.}
\setlength{\tabcolsep}{6pt}
{\fontsize{7}{10}\selectfont
\begin{tabular}{>{\raggedright\arraybackslash}p{3.2cm}|p{0.4cm}p{0.4cm}p{0.4cm}p{0.4cm}}
\toprule
Model & Short & Med & Long & Avg \\
\hline
\rowcolor{gray!20} \multicolumn{5}{l}{\textit{Closed-Source Models}} \\
Gemini-2.0-Flash &44.2 &\underline{60.3} &\underline{53.5} &\underline{51.7} \\
Gemini-1.5-Pro &37.4 &49.9 &48.7 &44.0 \\
o1 &\textbf{53.6} &\textbf{61.3} &\textbf{54.7} &\textbf{56.7} \\
GPT-4o &\underline{44.4} &48.7 &49.7 &46.9 \\
Claude 3.5 Sonnet &39.8 &42.2 &41.4 &41.0 \\
\midrule
\rowcolor{gray!20} \multicolumn{5}{l}{\textit{Open-Source Models}} \\
InternVL2.5-8B &20.7 &25.7 &28.3 &23.9 \\
InternVL2.5-78B &30.4 &30.5 &32.6 &30.9 \\
VideoLLaMA3-7B &30.2 &38.2 &26.7 &32.5 \\
LLaVA-OneVision-7B &29.2 &33.4 &28.9 &30.7 \\
LLaVA-OneVision-72B &35.1 &40.6 &31.0 &36.4 \\
mPLUG-Owl3-7B &6.1 &9.9 &4.8 &7.3 \\
MiniCPM-o2.6-8B &27.5 &26.0 &26.7 &26.9 \\
Llama-3.2-11B-Vision &5.3 &5.1 &3.7 &4.9 \\
Qwen2.5-VL-7B &27.1 &34.0 &31.6 &30.4 \\
Qwen2.5-VL-72B &33.4 &42.8 &39.8 &37.9 \\
LLaVA-Video-7B &31.7 &33.4 &32.6 &32.5 \\
LLaVA-Video-72B &35.5 &40.6 &38.5 &37.9 \\
Aria-25B &36.4 &39.9 &39.6 &38.2 \\
InternVideo2.5-8B &31.5 &35.0 &32.6 &33.0 \\
\bottomrule
\end{tabular}}
\label{tab:duration}
\end{minipage}
\hfill
\begin{minipage}{0.48\textwidth}

\centering
\captionof{table}{Accuracy Evaluation Results under Different Settings.}
\setlength{\tabcolsep}{6pt}
{\fontsize{7}{8}\selectfont
\begin{tabular}{>{\raggedright\arraybackslash}p{2.2cm}|>{\centering\arraybackslash}p{0.4cm}>{\centering\arraybackslash}p{0.8cm}>{\centering\arraybackslash}p{0.4cm}>{\centering\arraybackslash}p{0.4cm}}
\toprule
Model &Text&1 Frame& Direct& CoT \\
\hline
\rowcolor{gray!20} \multicolumn{5}{l}{\textit{Closed-Source Models}} \\
Gemini-2.0-Flash &\textbf{13.8} &\textbf{25.2} &\underline{44.8} & \textbf{51.7} \\
GPT-4o &9.8 &\underline{21.6} &\textbf{46.3} & \underline{46.9} \\
Claude 3.5 Sonnet &9.1 &11.3 &39.6 &41.0 \\
\midrule
\rowcolor{gray!20} \multicolumn{5}{l}{\textit{Open-Source Models}} \\
InternVL2.5-78B &7.2 &18.7 &35.4 & 30.9 \\
Qwen2.5-VL-72B &\underline{12.7} &16.7 &42.7 & 37.9 \\
\bottomrule
\end{tabular}}
\label{tab:setting}
\vspace{10pt}

\centering
\includegraphics[width=\linewidth]{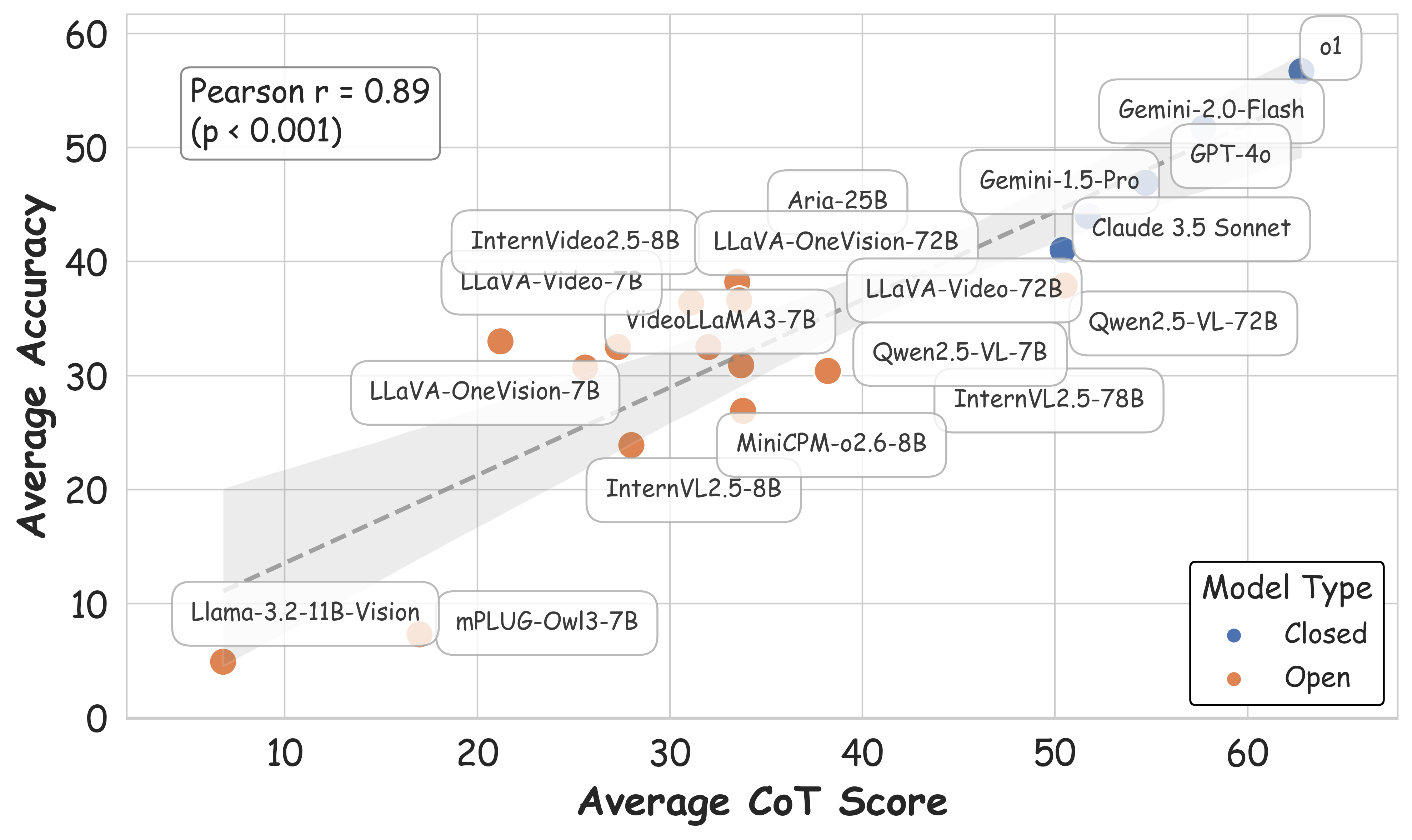}
\captionof{figure}{Correlation between CoT Evaluation Results and Accuracy Evaluation Results.}
\label{fig:correlation}

\end{minipage}
\end{figure}

%% file: section/5_conclusion.tex
\section{Conclusion}
We introduce VCR-Bench, the first benchmark specifically designed to evaluate the CoT reasoning capabilities of LVLMs in video understanding tasks. Our benchmark comprises a high-quality dataset of 859 videos and 1,034 QA pairs spanning seven distinct task types, each annotated with rigorous CoT reasoning references. We propose a novel evaluation framework that assesses reasoning quality through recall, precision, and their harmonic mean ($F_1$ score). Comprehensive evaluations reveal significant limitations in current LVLMs, with even the top-performing o1 model achieving only 62.8 CoT score and most open-source models scoring below 40, highlighting substantial room for improvement in video-grounded reasoning. VCR-Bench establishes a standardized framework to advance research in this critical area.

%% file: appendix/appendix.tex
\newpage
\appendix

\input{appendix/prompt}
\input{appendix/error_analysis}
\input{appendix/case}

%% file: appendix/prompt.tex
\newpage
\section{Prompt Template}
\label{appendix: prompt}

\begin{tcolorbox}[breakable, colback=gray!5!white, colframe=gray!75!black, 
title=Recall Evaluation Prompt, boxrule=0.5mm, width=\textwidth, arc=3mm, auto outer arc]
You are an expert system for verifying solutions to video-based problems. Your task is to match the ground truth middle steps with the provided solution.\\

\textbf{INPUT FORMAT:}

1. Problem: The original question/task

2. A Solution of a model

3. Ground Truth: Essential steps required for a correct answer \\

\textbf{MATCHING PROCESS:}

You need to match each ground truth middle step with the solution:\\

\textbf{Match Criteria:}\\
- The middle step should exactly match in the content or is directly entailed by a certain content in the solution\\
- All the details must be matched, including the specific value and content\\
- You should judge all the middle steps for whether there is a match in the solution\\

\textbf{Step Types:}\\
1. Logical Inference Steps\\
   \quad - Contains exactly one logical deduction\\
   \quad - Must produce a new derived conclusion\\
   \quad - Cannot be just a summary or observation\\

2. Video Description Steps\\
   \quad - Pure visual observations\\
   \quad - Only includes directly visible elements\\
   \quad - No inferences or assumptions\\
   \quad - Contains event time\\

\textbf{OUTPUT FORMAT:}\\
JSON array of judgments:
\begin{verbatim}
[
  {
    "step": ground truth middle step,
    "step_type": "Video Description Steps|Logical Inference Steps",
    "judgment": "Matched" | "Unmatched"
  }
]
\end{verbatim}

\textbf{ADDITIONAL RULES:}\\
1. Only output the json array with no additional information.\\
2. Judge each ground truth middle step in order without omitting any step.\\

Here is the problem, answer, solution, and the ground truth middle steps:

\textbf{[Problem]}: \{question\}

\textbf{[Answer]}: \{answer\}

\textbf{[Solution]}: \{solution\}
\end{tcolorbox}
\label{fig: prompt1}

\begin{tcolorbox}[breakable, colback=gray!5!white, colframe=gray!75!black, 
title=Precision Evaluation Prompt, boxrule=0.5mm, width=\textwidth, arc=3mm, auto outer arc]
Given a solution with multiple reasoning steps for a video-based problem, reformat it into well-structured steps and evaluate their correctness.\\

\textbf{Step 1: Reformatting the Solution}

Convert the unstructured solution into distinct reasoning steps while:\\
- Preserving all original content and order\\
- Not adding new interpretations\\
- Not omitting any steps\\

\textbf{Step Types}

1. Logical Inference Steps\\
   - Contains exactly one logical deduction\\
   - Must produce a new derived conclusion\\
   - Cannot be just a summary or observation\\

2. Video Description Steps\\
   - Pure visual observations\\
   - Only includes directly visible elements\\
   - No inferences or assumptions\\
   - Contains event time\\

3. Background Review Steps: \\
   - Repetition or review of the problem\\
   - Not directly related to solving the problem.\\

\textbf{Step Requirements}

- Each step must be atomic (one conclusion per step)\\
- No content duplication across steps\\
- Initial analysis counts as background information\\
- Final answer determination counts as logical inference\\

\textbf{Step 2: Evaluating Correctness}

Evaluate each step against:

\textbf{Ground Truth Matching}

For video descriptions:\\
- Key elements must match ground truth descriptions

For logical inferences:\\
- Conclusion must EXACTLY match or be DIRECTLY entailed by ground truth

For Background review: \\
-  Without special circumstances are deemed to be redundant\\

\textbf{Reasonableness Check (if no direct match)}

If Step:\\
- Premises must not contradict any ground truth or correct answer\\
- Logic is valid\\
- Conclusion must not contradict any ground truth\\
- Conclusion must support or be neutral to correct answer\\
- Helpful in solving the problem, non-redundant steps\\
this Step be viewed as matched.\\

\textbf{Judgement Categories}

- "Match": Aligns with ground truth\\
- "Wrong": Contradictory with ground truth\\
- "Redundant": Redundant steps that do not help solve the problem\\

\textbf{Output Requirements}

1. The output format MUST be in valid JSON format without ANY other content.\\
2. For highly repetitive patterns, output it as a single step.\\
3. Output maximum 35 steps. Always include the final step that contains the answer.\\

\textbf{Output Format}

\begin{verbatim}
[
  {
    "step": "reformatted the solution step",
    "step_type": "Video Description Steps|Logical Inference Steps|
                 Background Review Steps",
    "reasons_for_judgment": "The reason for judging...",
    "judgment": "Matched|Wrong|Redundant"
  }
]
\end{verbatim}

\textbf{Input Data}

\textbf{[Problem]:} \{question\}

\textbf{[Solution]: } \{solution\}

\textbf{[Ground Truth Information]:} \{gt\_annotation\}
\end{tcolorbox}
\label{fig:prompt1}

\begin{tcolorbox}[breakable, colback=gray!5!white, colframe=gray!75!black, 
title=Answer Extraction Prompt, boxrule=0.5mm, width=\textwidth, arc=3mm, auto outer arc]
You are an AI assistant who will help me to extract an answer of a question. You are provided with a question and a response, and you need to find the final answer of the question. \\

\textbf{Extract Rule:}

[Multiple choice question]

1. The answer could be answering the option letter or the value. You should directly output the choice letter of the answer. 

2. You should output a single uppercase character in A, B, C, D, E, F, G, H, I (if they are valid options), and Z.

3. If the answer is about a certain time period, such as from 1 minute 30 seconds to 2 minutes 30 seconds, it should be given in the format [90, 150].

4. If the meaning of all options are significantly different from the final answer, output Z. \\

[Non Multiple choice question]

1. Output the final value of the answer. It could be hidden inside the last step of calculation or inference. Pay attention to what the question is asking for to extract the value of the answer.

2. The final answer could also be a short phrase or sentence.

3. If the response doesn't give a final answer, output Z.\\

\textbf{Output Format:} 

Directly output the extracted answer of the response \\

\textbf{Example 1:}

Question: What is the main object in image?

Options: A. teddy bear B. rabbit C. cat D. dog

Response: a cute teddy bear

Your output: A\\

\textbf{Example 2: }

Question: What is the main object in image?

Options: A. teddy bear B. rabbit C. cat D. dog

Answer: Spider

Your output: Z\\

\textbf{Example 3:}

Question: 

On a flat playground, choose a Cartesian $Oxy$ coordinate system (in units of meters). A child running at a constant velocity $V = \SI{1}{\meter/\second}$ around a heart-shaped path satisfies the following order-6 algebraic equation:
\[
\left(x^{2} + y^{2} - L^{2}\right)^{3} - L x^{2} y^{3} = 0, \quad L = 10.
\]

Response: 

Given the radius of curvature \( r \approx \SI{10}{\meter} \):

Using the radius of curvature:
\[
a_c = \frac{V^2}{r} = \frac{(\SI{1}{\meter\per\second})^2}{\SI{10}{\meter}} = \SI{0.1}{\meter\per\second\squared}
\]

The magnitude of the acceleration is the sum of tangential and centripetal components. Since tangential acceleration is zero (constant speed):
\[
\text{Magnitude of acceleration} = a_c = \boxed{\SI{0.1}{\meter\per\second\squared}}
\]

Your output: $\SI{0.1}{\meter\per\second\squared}$ \\

\textbf{Example 4:}
Question: Given the sequence of images, please identify the image consistent with the text description: Clean And Jerk.

A. <image1>

B. <image2>

C. <image3>

D. <image4>,

Response: Based on the temporal analysis, Answer: The second image

Your output: B\\

\textbf{Example 5:}

Question: Please specify the start and end time of the event 'An intro screen appears with a gray background and the red and white words that say PRO TRAINING.'

Response: The start and end time of the event 'An intro screen appears with a gray background and the red and white words that say PRO TRAINING.' are as follows:

Step 1: Identify the start time of the event. 

The video starts at 00:00. The intro screen is visible from the very beginning. 

Start time: 00:00\

Step 2: Identify the end time of the event.

The intro screen starts to disappear and transition to the next scene at 00:11. By 00:12, the intro screen is completely gone. So, the intro screen is visible until 00:11.
End time: 00:11

Therefore, the event 'An intro screen appears with a gray background and the red and white words that say PRO TRAINING' starts at 00:00 and ends at 00:11.

Your output: [0,11]\\

\textbf{Example 6:}

Question: In the diagram, $O A=15, O P=9$ and $P B=4$. Determine the equation of the line through $A$ and $B$. Explain how you got your answer.

Response: The equation of the line through $A$ and $B$ is $y = -\frac{1}{3}x + 12$. 

To find the equation of the line through $A$ and $B$, we first need to find the slope of the line. The slope of a line passing through two points $(x_1, y_1)$ and $(x_2, y_2)$ is given by the formula:

$$\text{Slope} = \frac{y_2 - y_1}{x_2 - x_1}$$

Using the coordinates of points $A(0, 15)$ and $B(4, 0)$, we can calculate the slope as follows:

$$\text{Slope} = \frac{0 - 15}{4 - 0} = \frac{-15}{4} = -\frac{15}{4}$$

Now that we have the slope, we can use the point-slope form of a linear equation to find the equation of the line. The point-slope form is given by:

$$y - y_1 = m(x - x_1)$$

where $(x_1, y_1)$ is a point on the line and $m$ is the slope. In this case, we can use point $A(0, 15)$ and the slope $-\frac{15}{4}$:

$$y - 15 = -\frac{15}{4}(x - 0)$$

Simplifying the equation, we get:

$$y - 15 = -\frac{15}{4}x$$

$$y = -\frac{15}{4}x + 15$$

Therefore, the equation of the line through $A$ and $B$ is $y = -\frac{15}{4}x + 15$.

Your output: $y = -\frac{15}{4}x + 15$\\

\textbf{Example 7:}

Question: <image> <image> <image> <image> <image> <image> <image> <image> <image>In the paper `An Object is Worth 64x64 Pixels: Generating 3D Object via Image Diffusion', what aspect of the proposed method does this figure aim to demonstrate?

Response: the potential of this 3D generation paradigm to generate objects with PBR materials

Your output: the potential of this 3D generation paradigm to generate objects with PBR materials\\

\textbf{[Question]:} \{question\}

\textbf{[Answer]:} \{response\}

\textbf{[Your output]:}
\end{tcolorbox}
\label{fig: prompt3}

\begin{tcolorbox}[breakable, colback=gray!5!white, colframe=gray!75!black, 
title=Answer Scoring Prompt, boxrule=0.5mm, width=\textwidth, arc=3mm, auto outer arc]
You are an AI assistant who will help me to judge whether two answers are consistent. \\

\textbf{Input Illustration:}

[Standard Answer] is the standard answer to the question

[Model Answer] is the answer extracted from a model's output to this question. \\

\textbf{Task Illustration:}
Determine whether [Standard Answer] and [Model Answer] are consistent.

\textbf{Consistent Criteria:}

[Multiple-Choice questions]

1. If the [Model Answer] is the option letter, then it must completely matches the [Standard Answer].

2. If the [Model Answer] is not an option letter, then the [Model Answer] must completely match the option content of [Standard Answer].

[Nan-Multiple-Choice questions]

1. The [Model Answer] and [Standard Answer] should exactly match.

2. If the meaning is expressed in the same way, it is also considered consistent, for example, 0.5m and 50cm.

\textbf{Output Format: }

1. If they are consistent, output 1; if they are different, output 0.

2. DIRECTLY output 1 or 0 without any other content.\\

\textbf{Example 1:}

Question: What is the main object in image?

Options: A. teddy bear B. rabbit C. cat D. dog

[Model Answer]: a cute teddy bear

[Standard Answer]: A

Your output: 1\\

\textbf{Example 2: }

Question: Find the value of AB. 
Choices: A.1; B.5; C.9; D.10

[Model Answer]: 5

[Standard Answer]: B

Your output: 1\\

\textbf{Example 3: }

Question: Three of the following four slides are from the same presentation, but one is from a different one. Please identify the outlier: <image> <image> <image> <image>

[Model Answer]: the forth image

[Standard Answer]: the third image

Your output: 0\\

\textbf{[Question]: } \{question\}

\textbf{[Model Answer]:} \{extract answer\}

\textbf{[Standard Answer]:} \{gt answer\}

\textbf{Your output:}
\end{tcolorbox}
\label{fig: prompt4}

\clearpage

%% file: appendix/error_analysis.tex
\section{Error Analysis}
\label{appendix: error analysis}

\begin{figure*}[htbp]
\includegraphics[width=0.99\textwidth]{figure/error_case1.pdf} 
\caption{\textbf{Common error examples of the model.} The top image shows reasoning failures from missing critical visual features, while the bottom image demonstrates excessive inferences beyond the video content, leading to incorrect answers.}
\label{fig:error case}
\end{figure*}

\begin{figure*}[htbp]
\includegraphics[width=0.99\textwidth]{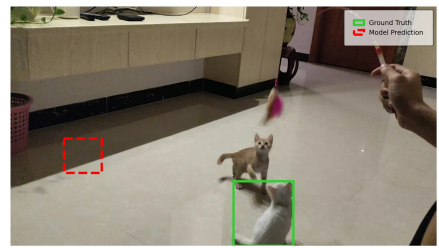} 
\caption{\textbf{Common error examples of the model.} The top example shows incorrect reasoning due to the model's lack of domain-specific knowledge, while the bottom one highlights significant deviations caused by poor spatiotemporal localization capabilities.}
\label{fig:error case2}
\end{figure*}

\begin{figure*}[htbp]
\includegraphics[width=0.99\textwidth]{figure/error_case3.pdf} 
\caption{\textbf{Common error examples of the model (Correct result, wrong process).} The top example shows the model incorrectly located the initial position of the object (cat) to be identified, while the bottom one shows the model mistakenly identified the person inside the car in the video as the standing person mentioned in the question.}
\label{fig:error case3}
\end{figure*}

%% file: appendix/case.tex
\newpage
\section{More Qualitative Examples}
\label{appendix: more qualitative examples}

\begin{figure*}[htbp]
\centering
\includegraphics[width=0.85\textwidth]{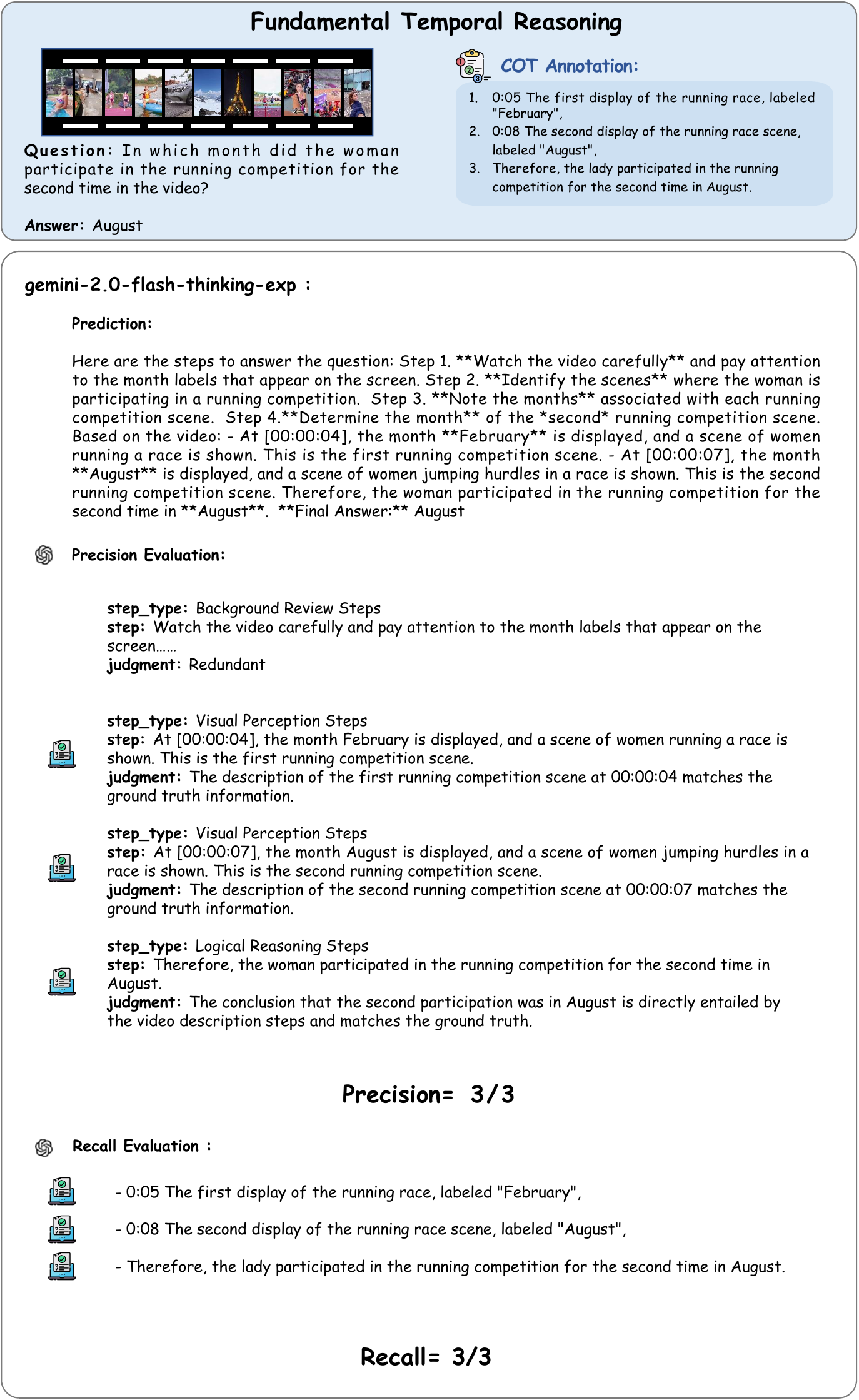} 
\caption{\textbf{Examples of FTR.}}
\label{fir:FTR_example_o1}
\end{figure*}
\newpage

\begin{figure*}[htbp]
\centering
\includegraphics[width=0.99\textwidth]{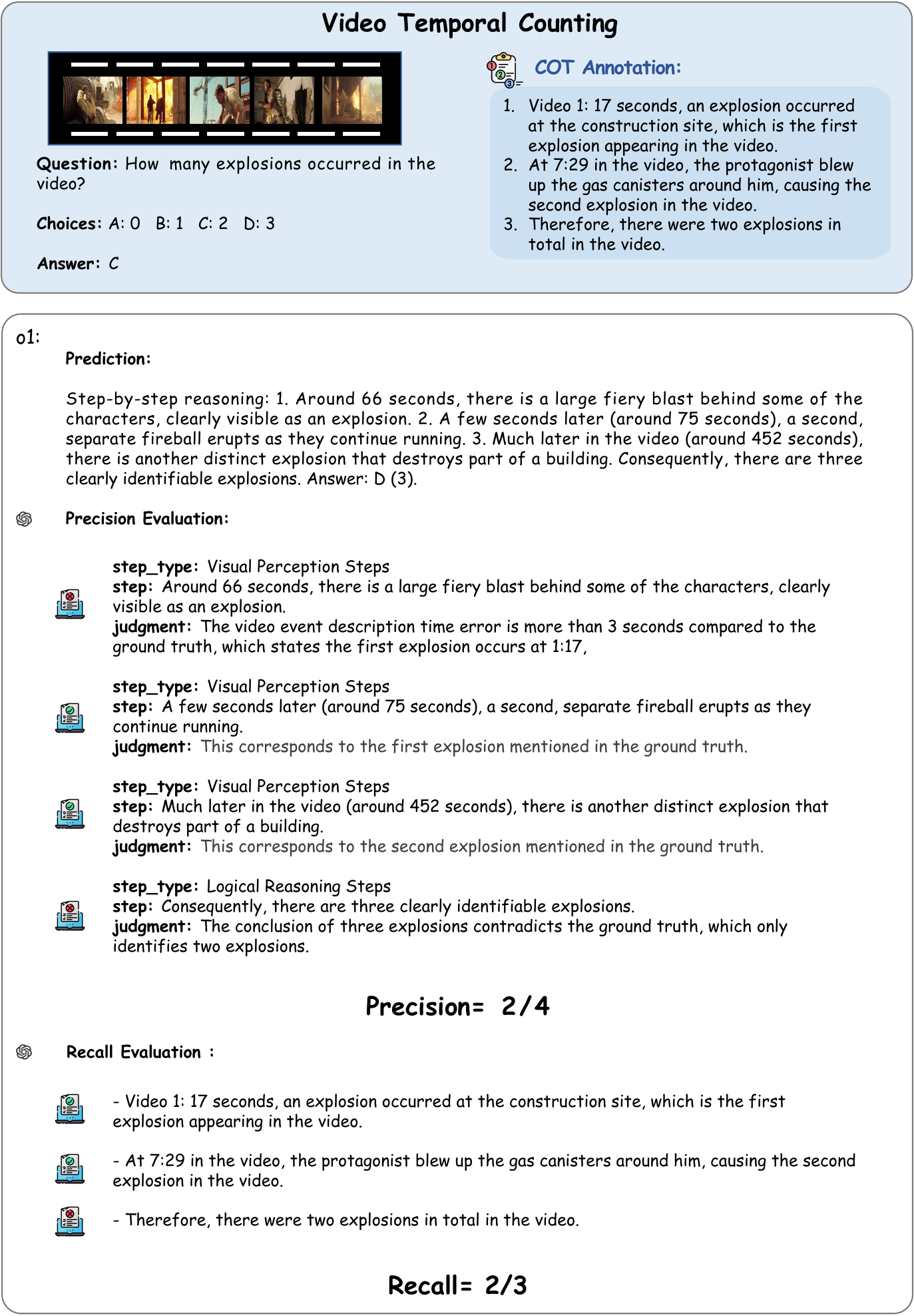} 
\caption{\textbf{Examples of VTC.}}
\label{fir:VTC_example_o1}
\end{figure*}
\newpage

\begin{figure*}[htbp]
\centering
\includegraphics[width=0.75\textwidth]{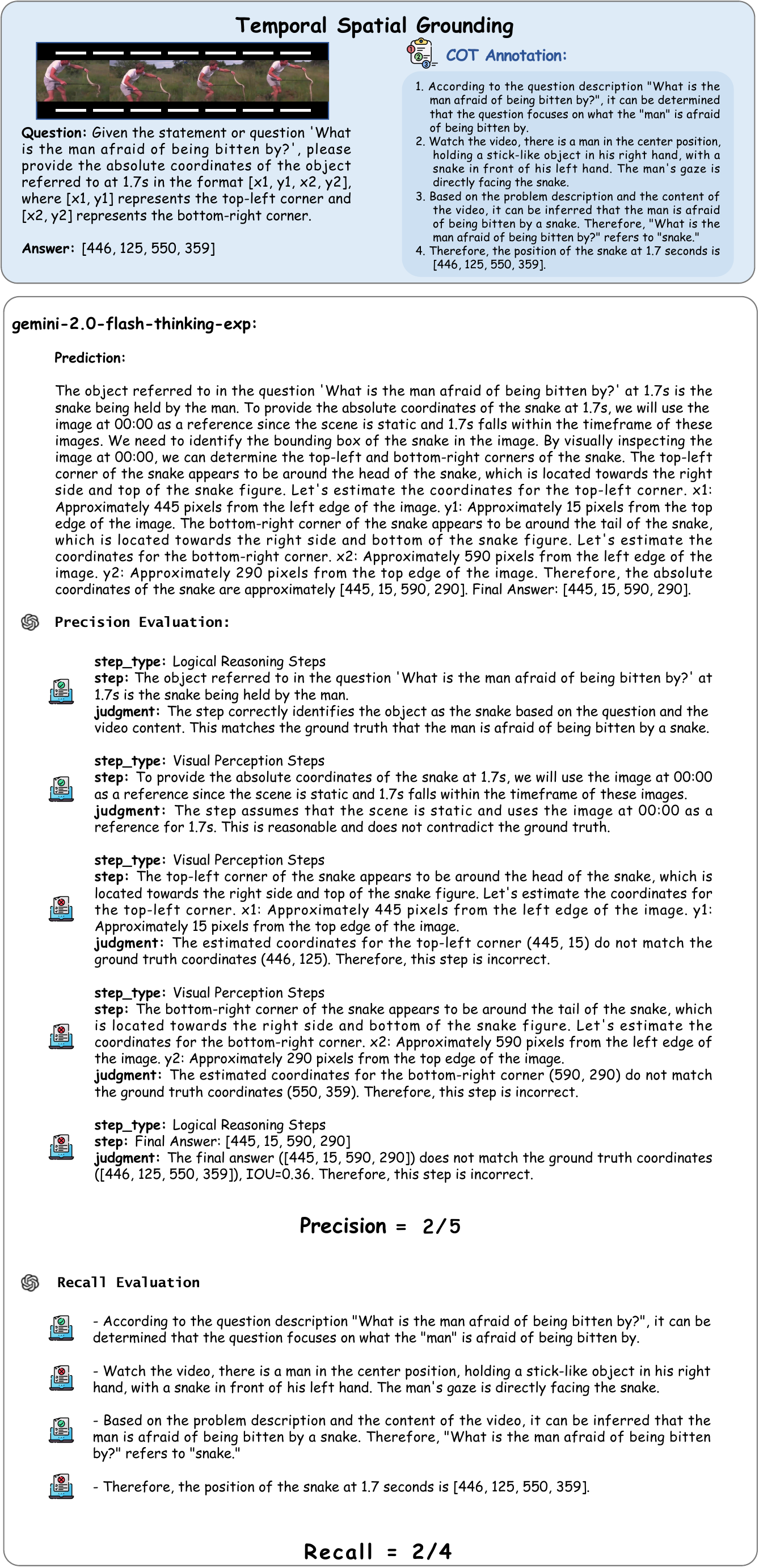} 
\caption{\textbf{Examples of TSG.}}
\label{fir:TSG_example_o1}
\end{figure*}
\newpage

\begin{figure*}[htbp]
\centering
\includegraphics[width=0.85\textwidth]{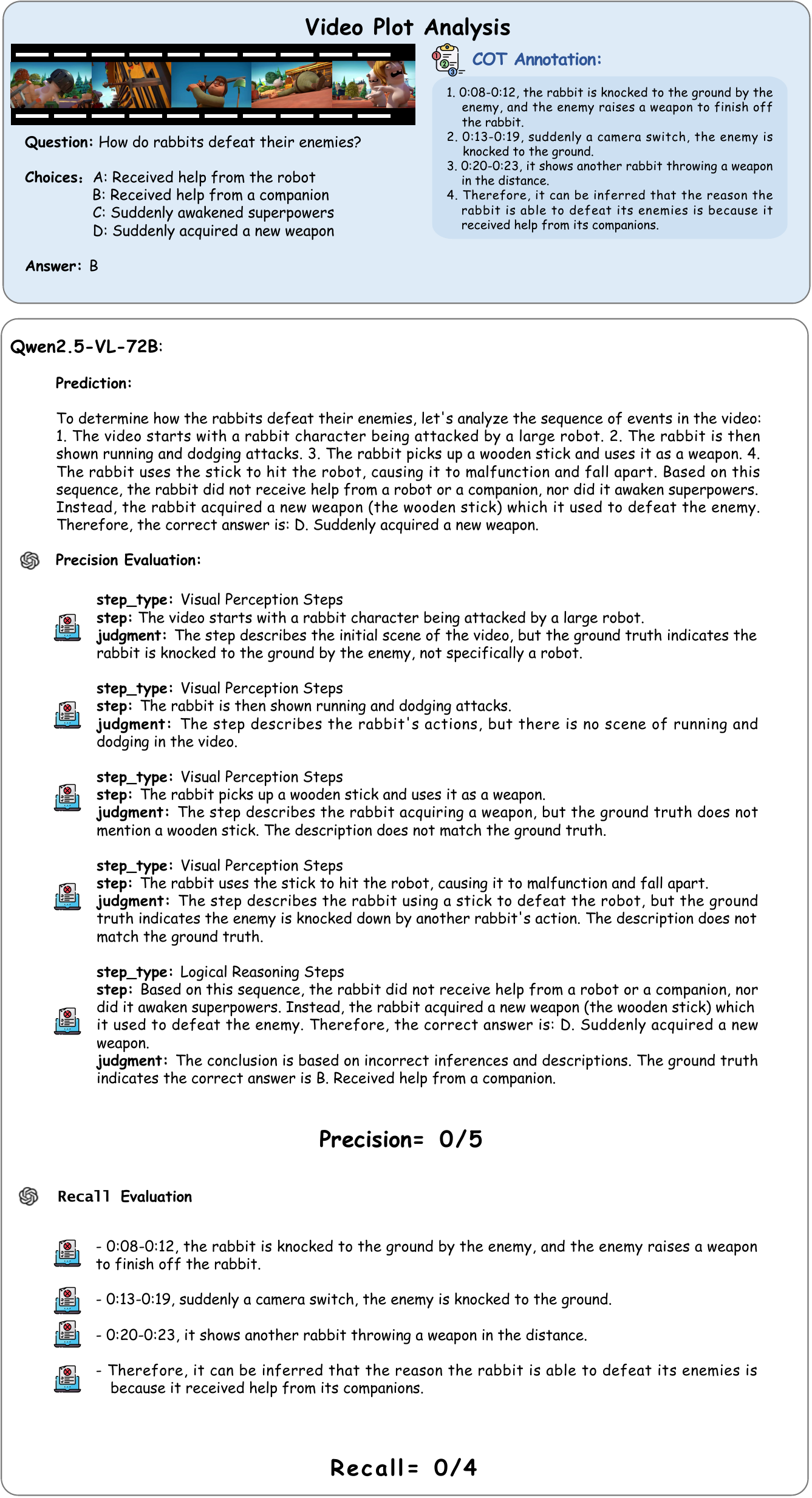} 
\caption{\textbf{Examples of VPA.}}
\label{fir:VPA_example_o1}
\end{figure*}
\newpage

\begin{figure*}[htbp]
\centering
\includegraphics[width=0.8\textwidth]{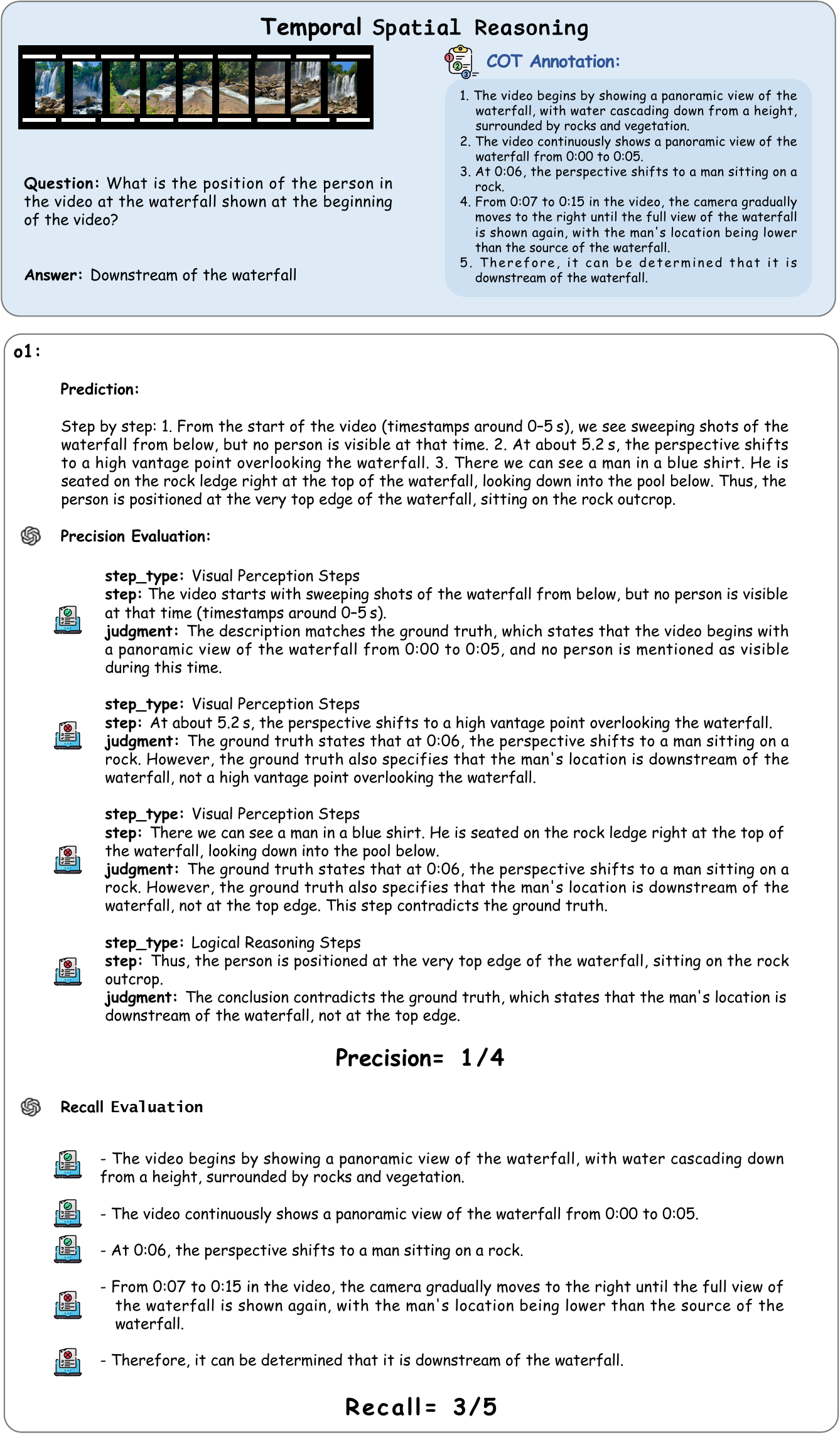} 
\caption{\textbf{Examples of TSR.}}
\label{fir:TSR_example_o1}
\end{figure*}
\newpage

\begin{figure*}[htbp]
\centering
\includegraphics[width=0.75\textwidth]{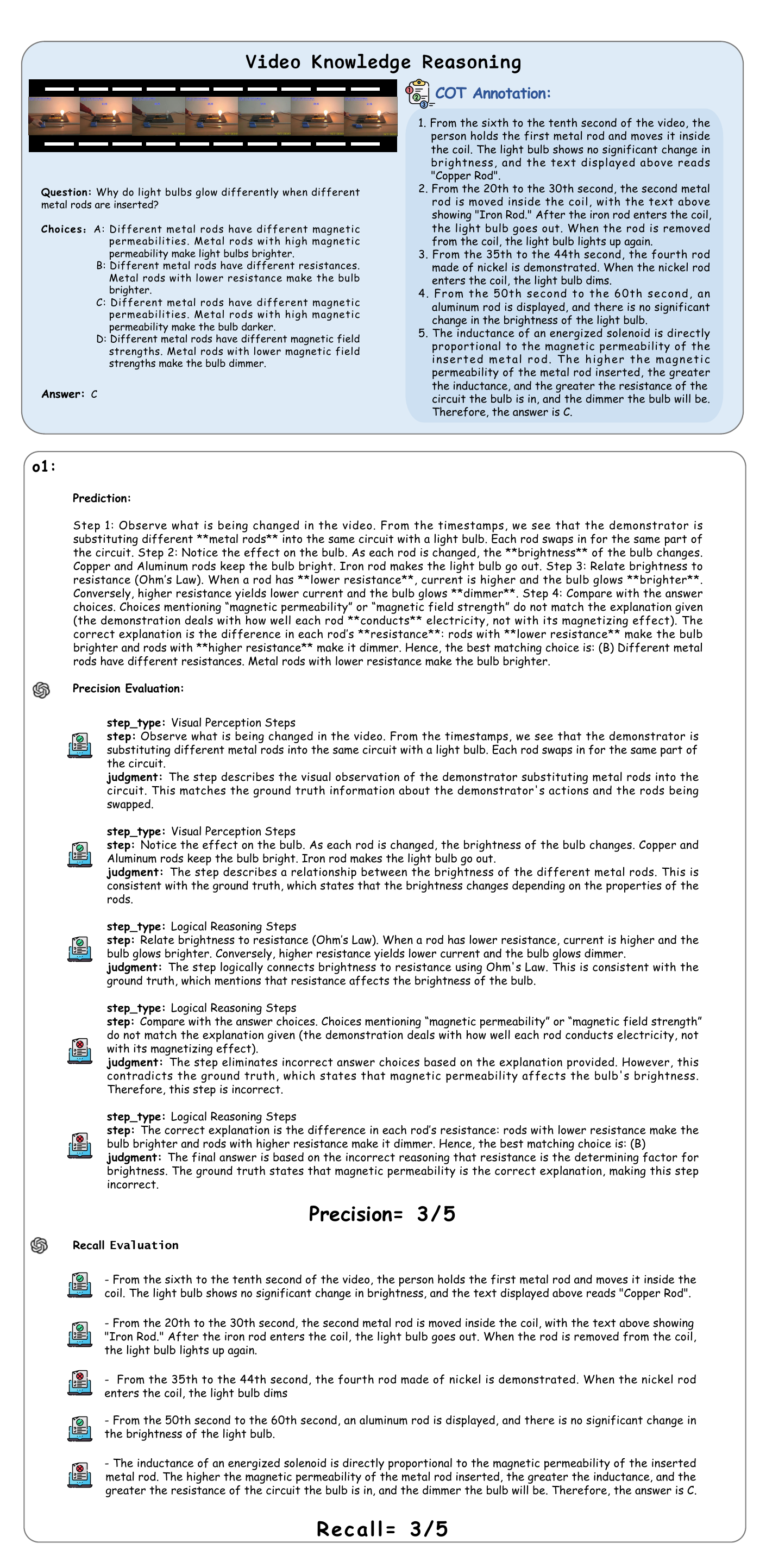} 
\caption{\textbf{Examples of VKR.}}
\label{fir:VKR_example_o1}
\end{figure*}
\newpage

\begin{figure*}[htbp]
\centering
\includegraphics[width=0.9\textwidth]{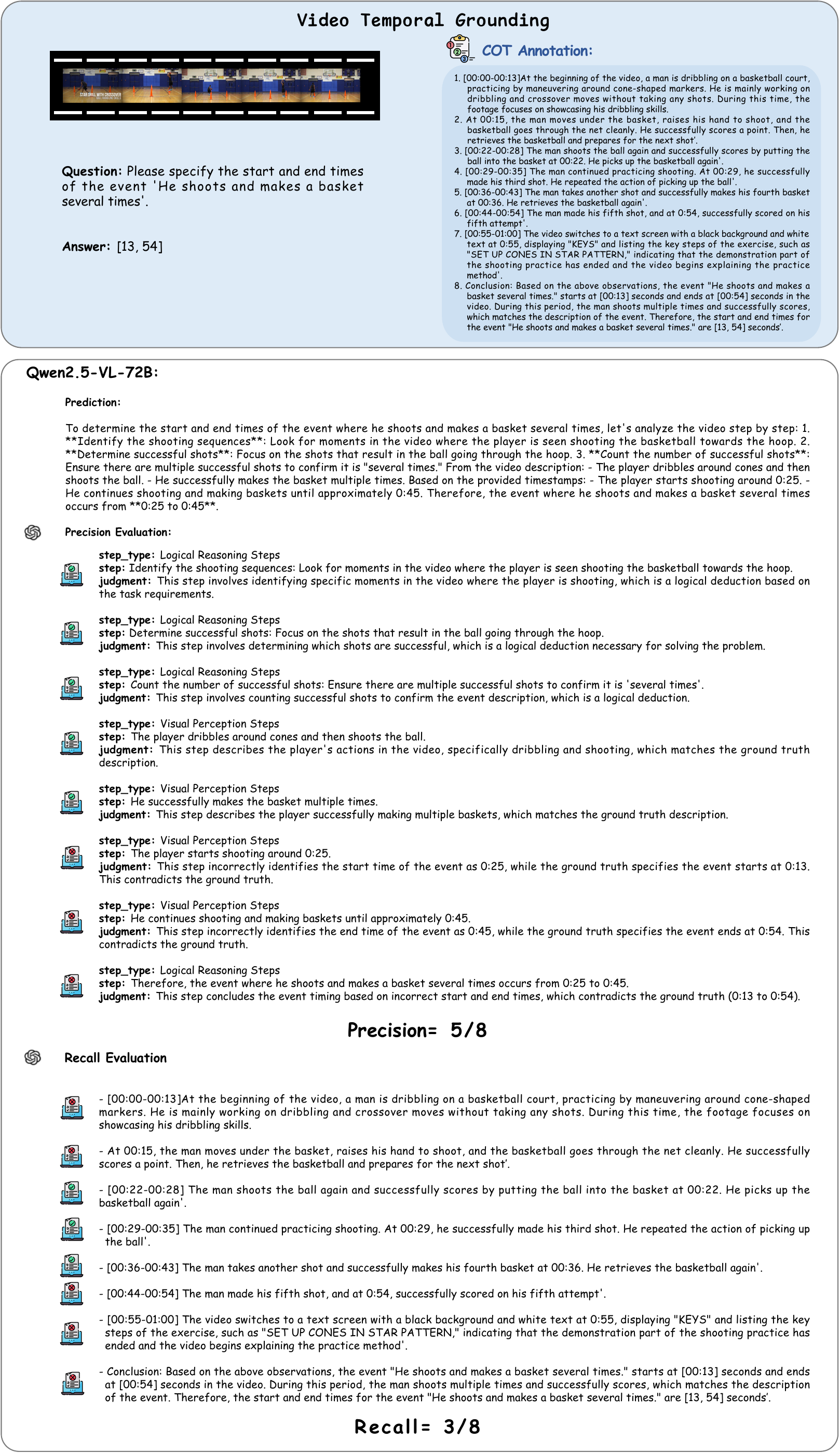} 
\caption{\textbf{Examples of VTG.}}
\label{fir:VTG_example_o1}
\end{figure*}
\newpage